\documentclass{article}
\usepackage[table]{xcolor}
\usepackage{hyperref}
\usepackage{enumerate}
\usepackage[utf8]{inputenc} 
\usepackage[T1]{fontenc}    
\usepackage{hyperref}       
\usepackage{url}            
\usepackage{booktabs}       
\usepackage{amsfonts}       
\usepackage{nicefrac}       
\usepackage{microtype}      
\usepackage{lipsum}
\usepackage{graphicx}
\usepackage{amssymb}
\usepackage{latexsym}
\usepackage{url}
\usepackage{xcolor}
\definecolor{newcolor}{rgb}{.8,.349,.1}
\usepackage{algorithm}
\usepackage{algorithmicx}
\usepackage[noend]{algpseudocode}
\usepackage[noend]{algpseudocode}
\usepackage{fancyhdr}
\usepackage{subfigure}
\usepackage[cmex10]{amsmath}
\usepackage{amsfonts}
\usepackage{wrapfig}
\usepackage{color}
\usepackage{verbatim}  
\usepackage{changepage} 
\usepackage{mdwlist}
\usepackage{paralist}
\usepackage{pdfpages}
\usepackage{graphicx}
\usepackage{float} 
\let\tabularnewline\\

\newcounter{defn}

\title{Machine Learning of polymer types from the spectral signature of Raman spectroscopy microplastics data}

\author{
	Sheela~Ramanna\thanks{Corresponding author. This research has been supported by the MITACS Grant IT-18982.} \\
	Department of Applied Computer Science\\ University of Winnipeg,\\
	Winnipeg, Manitoba R3B 2E9, Canada\\
	\texttt{s.ramanna@uwinnipeg.ca} \\
	$\And$
	Danila~Morozovskii\\
	Department of Applied Computer Science\\ University of Winnipeg,\\
	Winnipeg, Manitoba R3B 2E9, Canada\\
	\texttt{morozovskii-d@webmail.uwinnipeg.ca} \\
	$\And$
	Sam~Swanson\\
	Compound Connect\\ 
	Winnipeg, Manitoba Canada\\
	\texttt{204communications@gmail.com} \\
	$\And$
	Jennifer~Bruneau\\
	Compound Connect\\ 
	Winnipeg, Manitoba Canada\\
	\texttt{bruneau.jenn@gmail.com} \\
}

\begin{document}
\maketitle

\begin{abstract}
The tools and technology that are currently used to analyze chemical compound structures that identify polymer types in microplastics are not well-calibrated for environmentally weathered microplastics. Microplastics that have been degraded by environmental weathering factors can offer less analytic certainty than samples of microplastics that have not been exposed to weathering processes. Machine learning tools and techniques allow us to better calibrate the research tools for certainty in microplastics analysis. In this paper, we investigate whether the signatures (Raman shift values) are distinct enough such that well studied machine learning (ML) algorithms can learn to identify polymer types using a relatively small amount of labeled input data when the samples have not been impacted by environmental degradation. Several ML models were trained on a well-known repository, Spectral Libraries of Plastic Particles (SLOPP), that contain Raman shift and intensity results for a range of plastic particles, then tested on environmentally aged plastic particles (SloPP-E) consisting of 22 polymer types. After extensive preprocessing and augmentation, the trained random forest model was then tested on the SloPP-E dataset resulting in an improvement in classification accuracy of 93.81\% from 89\%.
\end{abstract}

\section{Introduction}
\label{sec:introduction}

Plastic pollution is exclusively the result of anthropogenic activities, with the majority of plastic entering the environment through land-based activities but ending up far from their source, having travelled though atmospheric and riverine pathways and degrading through multiple processes~\cite{Booth2020}. The durability and strength of plastics that make them suitable for a broad range of applications are also what cause them to disperse easily and have led to them becoming a global pollution problem. The primary reason they pose such a threat to the environment is their resistance to degradation, allowing them to persist for hundreds or thousands of years. However, their exposure to a variety of factors will result in them breaking down from macroplastics to microplastics~\cite{Conesa2020}.

Microplastics are composed of various polymers and include a broad array of chemical additives~\cite{Rochman2019}. It is understood that microplastics can decay at different rates depending on climate conditions, and that different stages of decay pose differing levels of toxicity to plant and animal life~\cite{Pflugmacher2020}. Thus, the chemical diversity of microplastics is an important consideration. The impacts of microplastics range from those on marine, freshwater, and terrestrial ecosystems~\cite{Booth2020}, on human health through ingestion of beverages and contamination in food and food packaging~\cite{Fournier2021}, and on microorganisms through uptake by zooplankton in freshwater ecosystems or interference with nutrient production and cycling in aquatic ecosystems. Finally, consumption of microplastics by humans through the food chain raises concerns about possible health risks and effects on the human body~\cite{Conesa2020}.

The two most promising techniques for microplastics (less than 5mm) analysis, are Raman and Fourier transform infrared (FTIR)
spectroscopy~\cite{Cabernard2018}. The preferred method for identifying microplastics is Raman spectroscopy which is an indispensable tool for the analysis of very small microplastics less than $20\mu$m~\cite{Araujo2018}. This is a vibrational spectroscopy technique based on the inelastic scattering of light~\cite{Zhu2019,Popov2020}. When laser light is shone on a plastic particle, a small amount of this light shifts in energy from the laser frequency because of interactions between the incident electromagnetic waves and the vibrational energy levels of the sample’s molecules. A Raman spectrum of the sample is created by plotting the Raman shift against the light frequency\footnote{\url{https://www.utsc.utoronto.ca/~traceslab/PDFs/raman_understanding.pdf}}. For example, in Fig~\ref{fig:slopp_example}, the y-axis gives the intensity of the scattered light, and the x-axis gives the energy of light. The specific type of material is marked with peaks in Raman spectroscopy.   One of the primary advantages of Raman spectroscopy is that even after exposure to ultraviolet (UV) light, the Raman spectra of microplastics are not significantly altered; this is significant as microplastics typically experience multiple forms of degradation with the majority of microplastics samples being degraded~\cite{Liu2020}. 

Machine learning (ML) algorithms such as Decision Trees (DT), Random Forest (RF), Support Vector Machines (SVM), K-Neighbour methods (KNN), Artificial Neural Networks (ANN) have been successfully applied to Raman spectra data in diverse areas of science. In~\cite{Madden2002}, ANN and KNN methods were used to predict the concentration of cocaine using Raman spectroscopy. The authors~\cite{Gniadecka2004}, use Raman spectroscopy to detect chemical changes in  melanoma tissue of patients and achieve  85\% (sensitivity) and 99\% (specificity) results with ANNs.  In~\cite{Pastor2018}, several well-known ML algorithms were applied to determine the mine of origin and extraction depth of samples by finding Raman spectral differences for variscite (phosphate mineral) specimens from the Gavà mining complex where the SVM classifier gave the best result of almost 90\% classification accuracy. In~\cite{KHAN2018}, the SVM model was able to achieve a diagnostic accuracy of 92\%  for tuberculosis patients using Raman spectra of blood sera.  RF classifier was used in the analysis of spectral information of various cultural heritage materials by~\cite{Sevetlidis2019}. The authors~\cite{Berghian-Grosan2020oils}, classify seven types of oils using Raman spectroscopy: sunflower, sesame, hemp, walnut, linseed (flaxseed), sea buckthorn and pumpkin seeds where a subspace KNN ensemble classifier gave the best classification accuracy of 88.9\%. In~\cite{Berghian-Grosan2020}, the authors explore association between Raman spectroscopy and machine learning to differentiate fruit distillate samples (alcoholic beverage) to determine trademark, geographical and botanical origin. The best geographical classification of the fruit distillates was obtained with the ensemble (subspace KNN) method resulting in an accuracy of 90.9\% for 30 samples. In~\cite{Ryzhikova2021}, ANN and SVM algorithms were used to diagnose biochemical composition of  biological fluids of patients with Alzheimer's disease based on near infrared (NIR) Raman spectroscopy with 84\% sensitivity and specificity values. The authors~\cite{Lussier2020}, present deep learning methods to extract and analyze chemical information in big and complex datasets derived from Raman and surface-enhanced Raman scattering (SERS) techniques. In~\cite{Houston2020}, 230 Raman spectra samples of high dimensional solvent and solvent mixtures (chemicals) were classified with deep neural networks using a locally connected architecture, resulting in a mean accuracy of 96.0\%.  In~\cite{Xia2020}, Raman spectra of oral tongue squamous cell carcinoma and para-carcinoma tissues of 24 patients were analyzed. A convolutional neural network model was used to extract features, which were then input to an SVM classifier resulting in a 99.96\% accuracy. AlexNet deep learning model was used to classify chronic renal failure using serum Raman spectra of 100 patients with an accuracy of 95.22\%~\cite{GAO2021}. 

In~\cite{Allen1999}, six types of common household plastics using Raman spectroscopy were evaluated to demonstrate the potential of machine learning methods such as principle component analysis, KNN as well regression models for classification and prediction. In~\cite{SHAN2018}, hyperspectral imaging was used to detect micro plastic contamination in soils. Classification precision of 86\% for polymers containining microplastics particles of size between 1-5 mm and about 99\% precision for microplastics particles of size between 0.5-1 mm  were obtained. In~\cite{Stefas2019} laser-induced breakdown spectroscopy was used to create plastic samples containing different additives such as flame retardants. Principle component analysis (PCA) and Linear Discriminant Analysis (LDA) were used to discriminate  11 different types of additives with LDA achieving almost 100\% accuracy. In~\cite{Agarwal2020}, 4000 images belonging to the five categories of plastic resin codes from a public database were classified using convolutional neural networks with an accuracy of 99.79\%.  In~\cite{Silva2020}, micro Fourier Transform Infrared ($\mu$-FTIR) hyper-spectral imaging with Partial least squares discriminant analysis (PLS-DA) and soft independent modelling of class analogy (SIMCA) which is based on PCA, were used to classify nine of the most common polymers in microplastics found on seabed sediment samples. A review of polymer informatics is presented in~ \cite{Sha2021}. In~\cite{HENRIKSEN2022}, PCA and clustering with K-means on short wave infrared hyperspectral data prepared using reflection imaging with a hyperspectral camera was used to analyze and classify 13 commercially available plastics. 

Our work differs from the more recent work where either hyperspectral imaging, digital images or laser-induced breakdown spectroscopy of plastics were used with machine learning models including deep learning models. The datasets used were either prepared by the authors or included large image repositories suitable for deep learning.   On the other hand, Raman spectroscopy data typically consist of approximately 1,000 to 3,000 data points. It is difficult and expensive to obtain the spectroscopy data, and only a limited amount of data is available online. Additionally, each sample might contain not one type of microplastic, but rather a combination of materials. To this end, in our work, several machine learning models were trained on a well-known repository, Spectral Libraries of Plastic Particles (SLOPP) containing 148 samples and 158 samples from Mendeley \footnote{https://data.mendeley.com/datasets/kpygrf9fg6/1}. The SLOPP library contains Raman shift and intensity results of Raman spectroscopy laboratory analyses conducted at the Rochman Lab \footnote{https://rochmanlab.wordpress.com/spectral-libraries-for-microplastics-research/} in the Department of Ecology and Evolutionary Biology at the University of Toronto for a range of plastic particles. This library also includes environmentally aged plastic particles (SloPP-E) containing 97 samples.   After extensive preprocessing and augmentation, the trained random forest model was then tested on the SloPP-E dataset resulting in an improvement in classification accuracy of 93.81\% from 89\%. This work contributes to the understanding of environmental polymers by validating the machine learning methods that improve the predictive capability of Raman spectroscopy data analysis.    

Our paper is organized as follows: In section~\ref{sec:datasets}, we give a description of the  open source spectroscopy datasets considered in this work. In section~\ref{sec:method}, we present a detailed discussion of the preprocessing and augmentation techniques used in this research for generating training and testing examples. In section~\ref{sec:res}, we give an in-depth analysis of the classification results of our final model followed by concluding remarks in section~\ref{sec:con}.
\section{MATERIALS- SPECTROSCOPY DATA}
\label{sec:datasets}
A Raman spectrum can provide molecular bond information on a particular substance and may be described as a “fingerprint” of the substance due to its uniqueness~\cite{Madden2008}. Raman spectra are a plot of scattered intensity as a function of the energy difference between the incident and scattered photons and are obtained by pointing a monochromatic laser beam at a sample~\cite{Movasaghi2007}. The resultant spectra are characterized by shifts in wave numbers (inverse of wavelength in $cm^-1$ ) from the incident frequency. The frequency difference between incident and Raman-scattered light is termed the Raman shift, which is unique for individual molecules. For this research, the following datasets have been used SLoPP, SLoPP-E, Mendeley.  A combined dataset of SLoPP and Mendeley was used as our training dataset, while SLoPP-E was used as the testing dataset. 
    
\begin{description}
	\item \textcolor{blue}{SLOPP:} SLoPP is a spectral library of microplastic particles with 148 samples, having different polymer types (shown in Table~\ref{tab:slopp_sloppe}), colours and morphologies.  Examples of colours are turqoise, orange, green, white, grey, black, light brown and clear. Examples of morphologies include: fragments, sphere, film, foam, and fiber.
SLoPP was collected in the range of 100-3500 $cm^-1$ and was created to include commonly used plastics.

Fig.~\ref{fig:slopp_example} illustrates the Raman spectra for one type of polymer (polypropylene).
The y-axis shows the intensity of the scattered light, and the x-axis shows the energy (frequency) of light. Different colours represent different samples in the dataset. It can be observed, that the most distinguishing feature of the Raman spectroscopy is spikes on different energies of the light.
	
	\begin{figure}[h!]
    	\includegraphics[width=12cm]{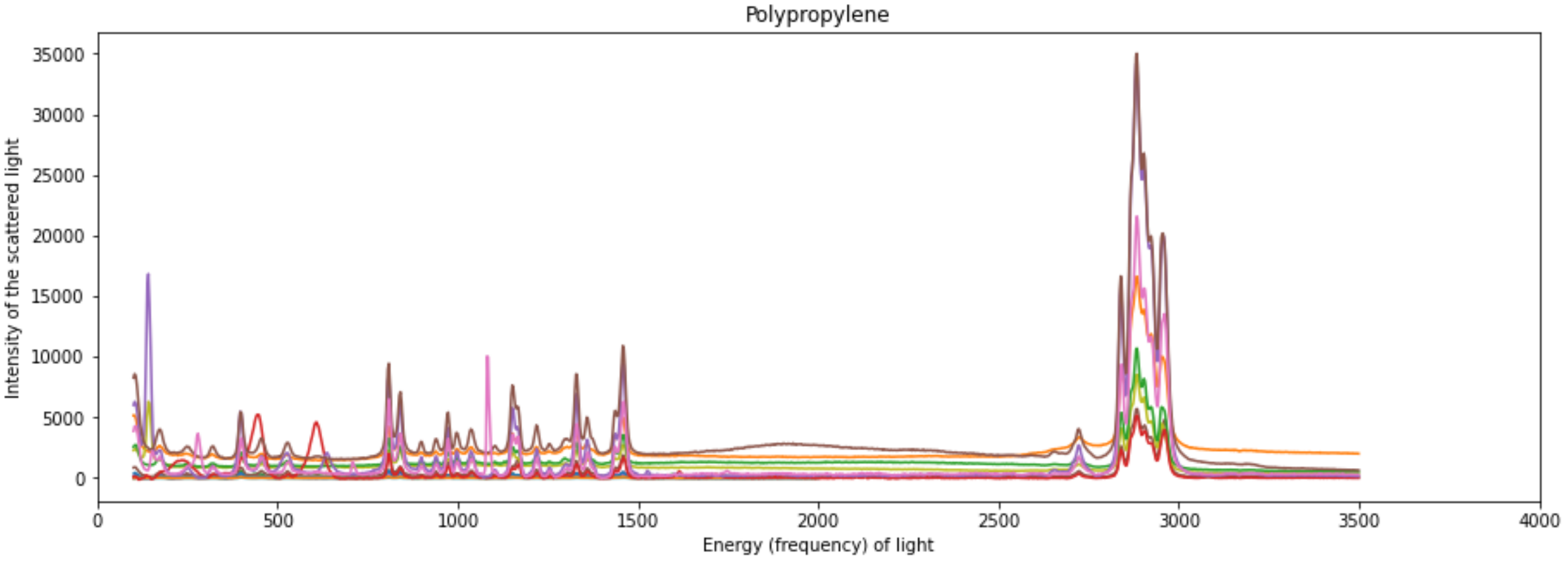}
    	\centering
    	\caption{Raman spectra of \textcolor{blue}{Polypropylene} from the SLoPP dataset.}
    	\label{fig:slopp_example}
    \end{figure}
	
	\item \textcolor{blue}{SLOPP-E:}  SLoPP-E dataset is similar to the SLoPP dataset, however, it includes samples exposed to a variety of environmental conditions (e.g., some samples have undergone some chemical degradation, ageing). The microplastics in this library SLoPP-E include environmental samples obtained across a range of matrices, geographies, and time. Fig. ~\ref{fig:sloppe_example} llustrates the Raman spectra for the same type of polymer (polypropylene) as the one shown in Fig.~\ref{fig:slopp_example}. Different colours represent different samples in the dataset. It can be observed that these two datasets share the same values on the x-axis (frequency) and similar intensities (spikes on the y-axis) for  the same polymer type. 
	 
	\begin{figure}[h!]
    	\includegraphics[width=12cm]{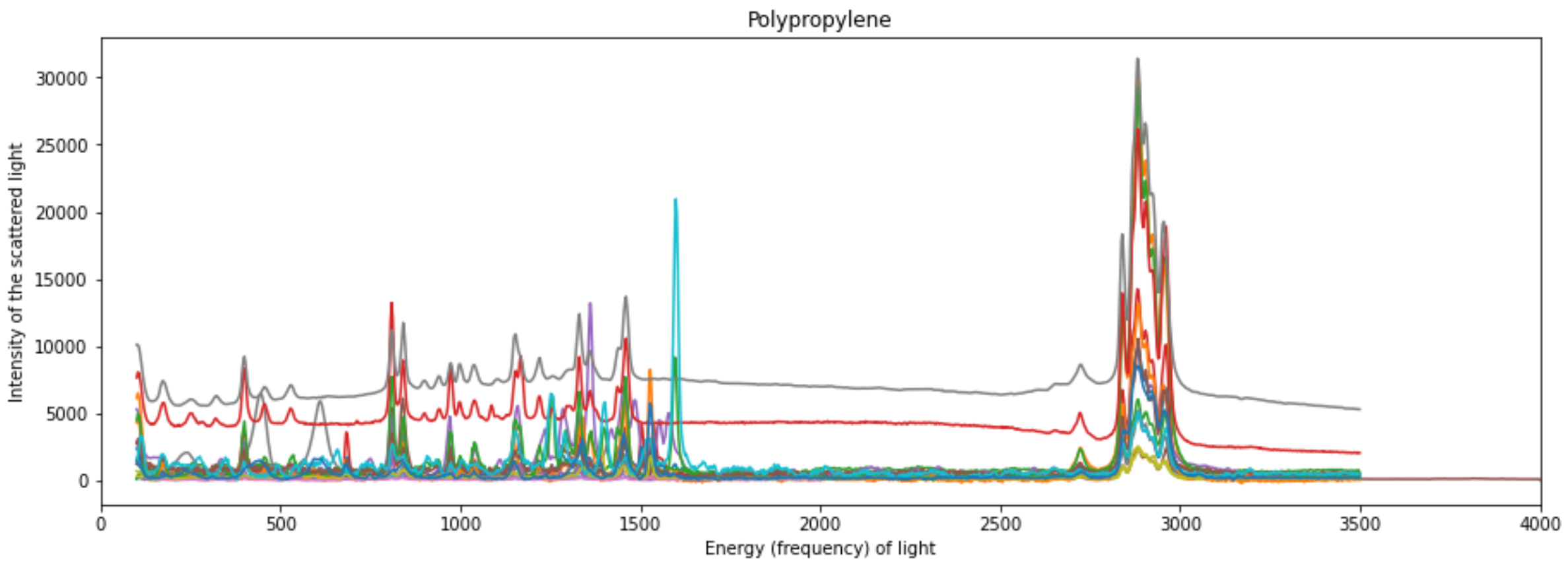}
    	\centering
    	\caption{Raman spectra of \textcolor{blue}{Polypropylene} from the SLoPP-E dataset.}
    	\label{fig:sloppe_example}
    \end{figure}
    
	\end{description}
    
Table~\ref{tab:slopp_sloppe} shows the distribution of polymers types for SLoPP and SLoPP-E. It can seen that some types are either missing in training (SLoPP) or testing (SLoPP-E) datasets. However, the training dataset includes many more types that are missing compared to the testing dataset. 

	\begin{center}
\begin{table}[h!]
\centering
		\begin{tabular}{|l|l|l|}
		\hline
            \textcolor{blue}{Polymer Types}                    & \textcolor{blue}{SLoPP samples} & \textcolor{blue}{SLoPP-E samples} \\
						\hline
            Acrylic                                     & 10             & 3                \\
            Acrylonitrile Butadiene Styrene             & 10             & 1                \\
            Cellulose Acetate                           & 4              & 3                \\
            Cotton                                      & 16             & -                \\
            Polyamide                                   & 7              & 7                \\
            Polycarbonate                               & 7              & 2                \\
            Polyester                                   & 10             & 12               \\
            Polyethylene                                & 24             & 26               \\
            Polyethylene Terephthalate                  & 9              & 1                \\
            Polyethylene Vinyl Acetate                  & 5              & -                \\
            Polymethyl Methacrylate                     & 1              & 3                \\
            Polypropylene                               & 17             & 21               \\
            Polystyrene                                 & 11             & 9                \\
            Polyurethane                                & 6              & 6                \\
            Polyvinyl Chloride                          & 11             & 3                \\
            Dyed Cellulose                              & -              & 5                \\
            Polybutylene Terephthalate                  & -              & 1                \\
            Polyethylene Terephthalate-co-Polycarbonate & -              & 1                \\
            Polyethylene-co-Polypropylene               & -              & 3                \\
            Polystyrene-co-Polyvinyl Chloride           & -              & 1                \\
            Polysulfone                                 & -              & 1                \\
            Rubber                                      & -              & 4         \\
									\hline
            \end{tabular}
        \caption{Data Distribution for SLoPP and SLoPP-E.}
        \label{tab:slopp_sloppe}
    \end{table}
    \end{center}
		
\begin{description}	
	\item \textcolor{blue} {Mendeley:} This dataset has two variations of microplastics: standard and weathered. The standard data is similar to SLoPP and the weathered data is similar to SLoPP-E (by description), subjected to environmental conditions. 
	
		\begin{figure}[h!]
    	\includegraphics[width=12cm]{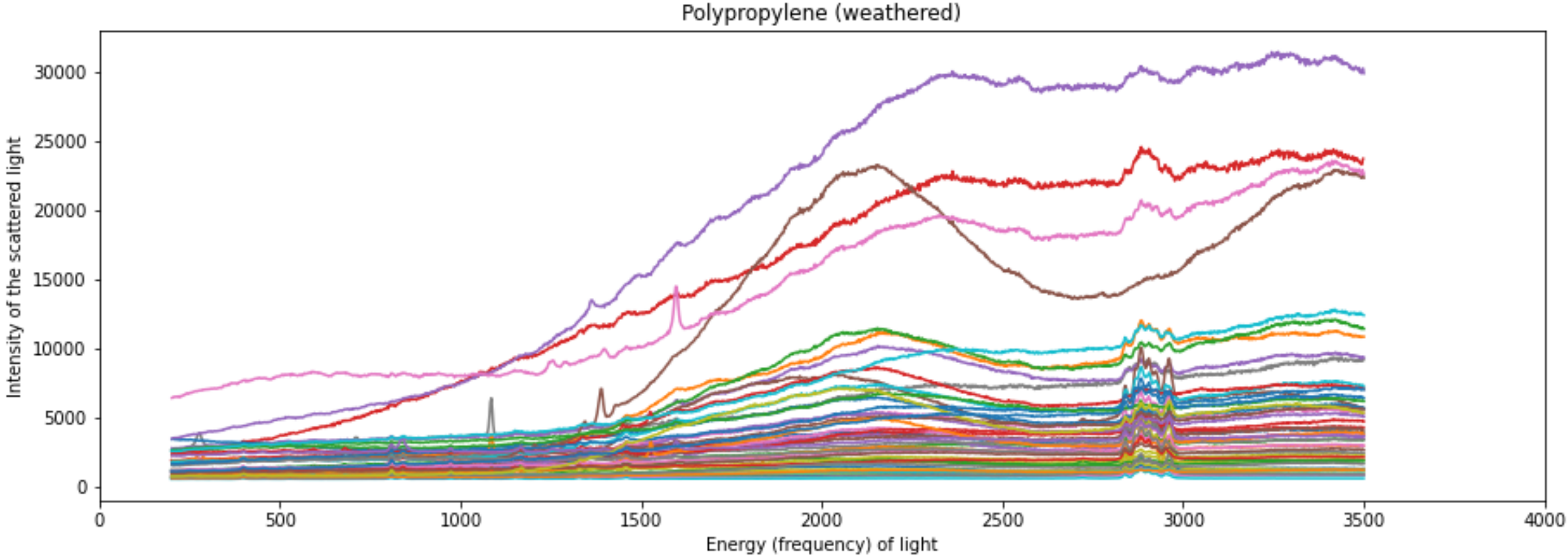}
    	\centering
    	\caption{Raman spectra of \textcolor{blue}{Polypropylene} from the Mendeley dataset.}
    	\label{fig:mendeley_example}
    \end{figure}
			
A plot of the Raman spectroscopy for polypropylene is shown in Fig.~\ref{fig:mendeley_example}. From this plot, one can observe the following : i) some samples in the Mendeley dataset have a wave-like structure, and ii)  the peaks (intensities of scattered light) are not as sharp and separated as in the SLoPP or SLoPP-E datasets. Table~\ref{tab:mendeley} shows the data distribution for Mendeley dataset. The majority of samples in this dataset belong to two polymer types: polypropylene and polyethylene and can be used in the training dataset, as they are also present in the SLoPP dataset.

	\begin{center}
\begin{table}[h!]
\centering
		\begin{tabular}{|l|l|}
		\hline
            \textcolor{blue}{Polymer Types}        & \textcolor{blue}{Mendeley samples} \\
						\hline
            Not detected                    & 8                 \\
            Acrylonitrile Butadiene Styrene & 1                 \\
            Nitrocellulose                  & 1                 \\
            Polyamine (nylon)               & 6                 \\
            Polycarbonate                   & 2                 \\
            Polyethylene                    & 74                \\
            Polyester                       & 16                \\
            Polypropylene                   & 54                \\
            Polystyrene (maybe)             & 2                 \\
            Polyvinyl chloride              & 9   \\
												\hline
        \end{tabular}
        \caption{Data Distribution for Mendeley.}
        \label{tab:mendeley}
    \end{table}
		\end{center}

		\end{description}

\subsection{Final dataset}
The final dataset used in our experiments is shown in Table \ref{tab:combined_dataset}. Note, that only the polymer types that are present in SLoPP are used. The majority of samples come from SLoPP, however, Mendeley contains a lot of samples for the polyester, polyethylene and polypropylene polymer types. The test dataset consists of only SLoPP-E, which was reduced to match the classes  (SLOPP polymer types) present in the training set. 16 samples from 7 different types of plastic have been removed resulting in a combined dataset of \textcolor{blue}{306} training samples and \textcolor{blue}{97} testing samples.

\begin{center}
\begin{table}[h!]
\centering
		\begin{tabular}{|l|l|l|l|}
		\hline
        \textcolor{blue}{Polymer Types}        & \textcolor{blue}{SLoPP} & \textcolor{blue}{SLoPP-E} & \textcolor{blue}{Mendeley} \\
				\hline
        Acrylic                         & 10             & 3       & -        \\
        Acrylonitrile Butadiene Styrene & 10             & 1       & 1        \\
        Cellulose Acetate               & 4              & 3       & -        \\
        Cotton                          & 16             & -       & -        \\
        Polyamide                       & 7              & 7       & -        \\
        Polycarbonate                   & 7              & 2       & 2        \\
        Polyester                       & 10             & 12      & 16       \\
        Polyethylene                    & 24             & 26      & 74       \\
        Polyethylene Terephthalate      & 9              & 1       & -        \\
        Polyethylene Vinyl Acetate      & 5              & -       & -        \\
        Polymethyl Methacrylate         & 1              & 3       & -        \\
        Polypropylene                   & 17             & 21      & 54       \\
        Polystyrene                     & 11             & 9       & 2        \\
        Polyurethane                    & 6              & 6       & -        \\
        Polyvinyl Chloride              & 11             & 3       & 9       \\
				\hline
    \end{tabular}
    \caption{Final dataset (SLoPP, SLoPP-E, Mendeley).}
    \label{tab:combined_dataset}
\end{table}
\end{center}

\section{METHODS: FEATURE ENGINEERING AND PREPROCESSING}
\label{sec:method}

As has been discussed in the previous section, different polymer types can be identified by the location of spikes on the x-axis (energy). Before this data can be used for classification learning, feature engineering which includes data transformation as well as preprocessing techniques such as normalization and discretization have been used. These techniques are described below.

\subsection{Normalization}
\label{sec:norm}
Here we discuss scaling methods for normalizing both the intensity (y-axis) and energy (x-axis) feature values since there are multiple problems with the feature values such as: varying ranges, varying step values, integer vs. real values as well negative values.
Alg.~\ref{alg:scaling} gives the pseudo-code for scaling the energy values.

\begin{itemize}

	\item \textcolor{blue}{Energy values}: Each sample in the dataset has a different x-axis range (i.e., one sample might have y-axis values between 100 and 1200 on the x-axis and another one between 300 and 3000). Furthermore, each sample's range between individual points on the x-axis is different as well (i.e., one sample can have a step value of 2 and another sample with a step value of 3). Therefore, scaling of the x-axis should be performed, where all values would be mapped to the corresponding points. Additionally, x-axis values are continuous values (ex: real value of 101.23), therefore, x-axis values should be mapped to integer values.

Scaling works as follows: firstly, as each sample has a different x-axis range, these values are mapped to the same range, by finding the minimum and maximum value of x for all samples (shown as parameter \textcolor{blue}{min\_range, max\_range} in Alg.~\ref{alg:scaling}).  In the case of the combined dataset, these parameter values are set to 0 and 3500 respectively. Then the values are populated by either the first value if the values occur at the beginning of the dataset, or by the last value if the values occur at the end. For example, if a sample has values on the x-axis ranging between 100 and 3000, then the values between 0-99 are populated with 100, and the values  between 3001-3500 are populated with the value 3000.

\begin{algorithm}
\caption{Scaling(Dataset, min\_range, max\_range)}\label{alg:scaling}
\begin{algorithmic}[1]

\State $dataset\_scaled \gets dict()$

\For{$plastic\_type$, $idxs$ in $Dataset.items()$}

    \State $dataset\_scaled[plastic\_type] \gets []$

    \For{$idx$ in $idxs$}
    
        \State $changed\_data \gets DataFrame(\{'x': range(min\_range, max\_range + 1),$ $'y': [0.0] * (max\_range + 1)\})$
        
        \State $last\_element\_idx \gets -1$
        
        \For{$index$, $point$ in $idx.iterrows()$}
        
            \State $idx\_of\_el \gets int(point['x'])$
        
            \If{$idx\_of\_el > max\_range$} 
                \State break
            \EndIf
        
            \If{$last\_element\_idx$ != $-1$} 
                \For{$i$ in $range(last\_element\_idx + 1, idx\_of\_el)$}
                    \State $changed\_data.at[i,$ $'y'] \gets changed\_data.at[last\_element\_idx,$ $'y']$
                \EndFor
            \Else
                \For{$i$ in $range(idx\_of\_el)$}
                    \State $changed\_data.at[i,$ $'y'] \gets point['y']$
                \EndFor
            \EndIf
            
            \State $changed\_data.at[idx\_of\_el,$ $'y'] \gets point['y']$
            \State $last\_element\_idx \gets idx\_of\_el$
            
        \EndFor
        
        \For{$i$ in $range(last\_element\_idx + 1, max\_range + 1)$}
            \State $changed\_data.at[i,$ $'y'] \gets changed\_data.at[last\_element\_idx,$ $'y']$
        \EndFor
        
        \State $dataset\_scaled[plastic\_type] \gets dataset\_scaled[plastic\_type] + changed\_data$
        
    \EndFor

\EndFor

\State return $dataset\_scaled$

\end{algorithmic}
\end{algorithm}

Secondly, as samples have a different step between each value, the gaps between these values are populated with the value which is at the beginning of a gap (i.e., if the x-values of two samples are 100 and 103, then all x-values having either 101 and 102 are replaced with value 100). As a result, this function produces 3501 points (3500 - 0 + 1), which are populated using the information from the original sample. 

\end{itemize}

\begin{itemize}

	\item  \textcolor{blue}{Intensity values}:	Some samples in the SLoPP-E test set have negative values for the intensity (y-axis).  Hence, all values have been scaled by adding a constant factor of one unit, which is the minimum negative value on the y-axis plus 1. This also ensures that there are no zero values.
	
\end{itemize}

\subsection{Data Transformation}
\label{sec:roc}

Two well-known data transformation techniques were used: Rate of Change (ROC) and Percentage Change (PC) shown in Eqns.~\ref{eq:roc} and \ref{eq:pc}. Both these techniques modify the original data by making sharp changes in the original dataset more visible.  

\begin{itemize}
	\item \textcolor{blue}{Rate of Change (ROC)}:  
\begin{equation}
\label{eq:roc}	
ROC = \frac{f(b)-f(a)}{b-a} 
\end {equation}	
\end{itemize}

\noindent where f(a) and f(b) are values on the y-axis and a and b are their corresponding values on the x-axis.

\begin{itemize}
	\item  \textcolor{blue}{Percentage Change (PC)}:	
	\begin{equation}
	\label{eq:pc}
	PC = \frac{f(a)}{mean (f(a-1),.... ,f(a-n))} 
	\end {equation}
	
\end{itemize}

\noindent where f(a) is the current value of the intensity (y-axis) and n is the number of values on the y-axis.

\begin{figure}[h!]
	\includegraphics[width=12cm]{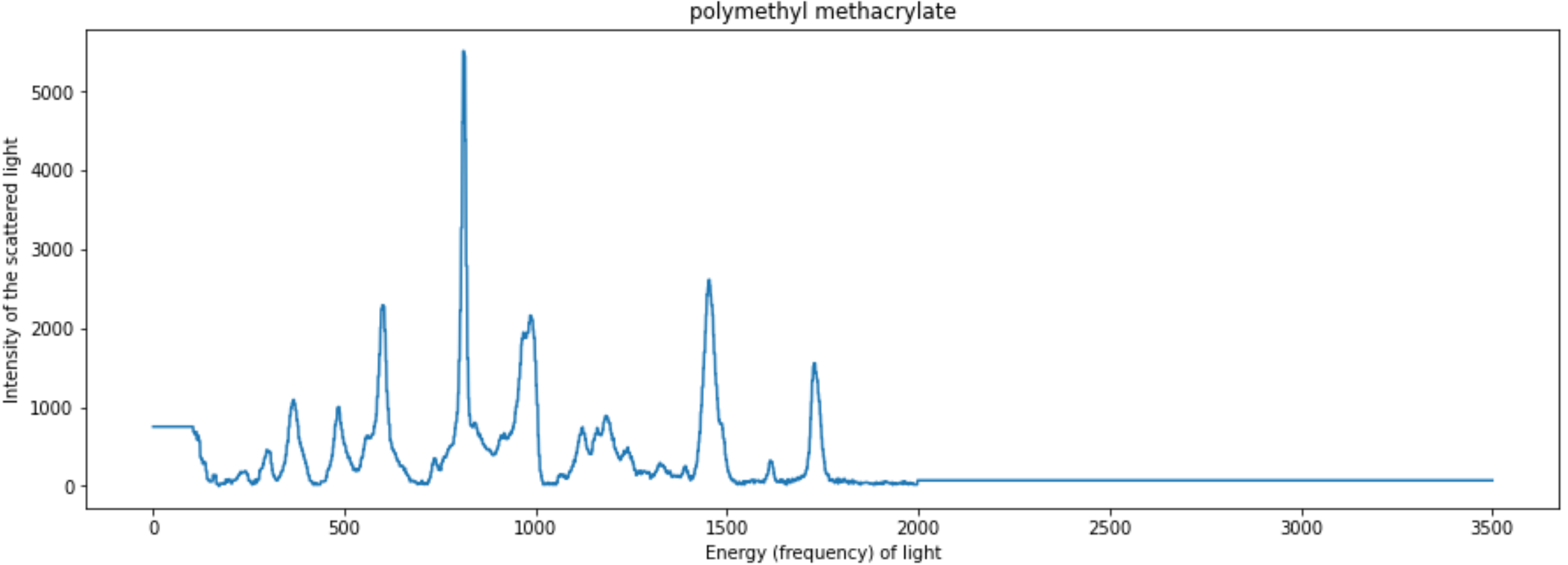}
	\centering
	\caption{Sample for polymethyl methacrylate.}
	\label{fig:roc_original}
\end{figure}

\begin{figure}[h!]
	\includegraphics[width=12cm]{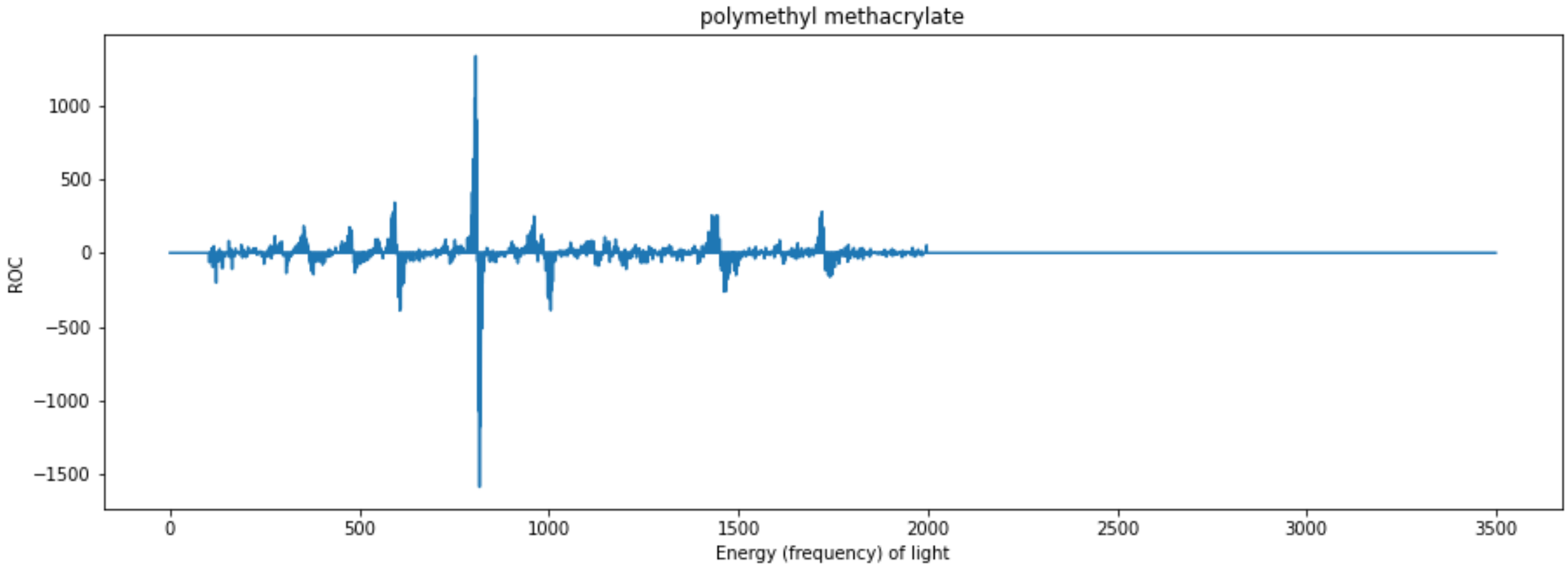}
	\centering
	\caption{ROC processed sample for polymethyl methacrylate.}
	\label{fig:roc_changed}
\end{figure}

Since the PC function did not give good classification results, the ROC function was used as the main data transformation technique. However, the PC function was used in the augmentation of the training set, which is described later in Sec.~\ref{sec:augment}. As an illustration of this technique, we present two figures. Fig.~\ref{fig:roc_original} shows the plot for a single sample of type polymethyl methacrylate.  Fig.~\ref{fig:roc_changed} shows the transformed plot.   The ROC transformation was applied to the intensity values (y-axis) which results in sharp spikes and preserves the changes in intensity values at the same energy (x-axis) co-ordinate.  It should be noted that, the values on the y-axis can be either positive or negative, meaning a positive or negative rate of change.

\subsection{Discretization- smoothing by bin-means}
\label{sec:discrete}

\begin{figure}[h!]
	\includegraphics[width=12cm]{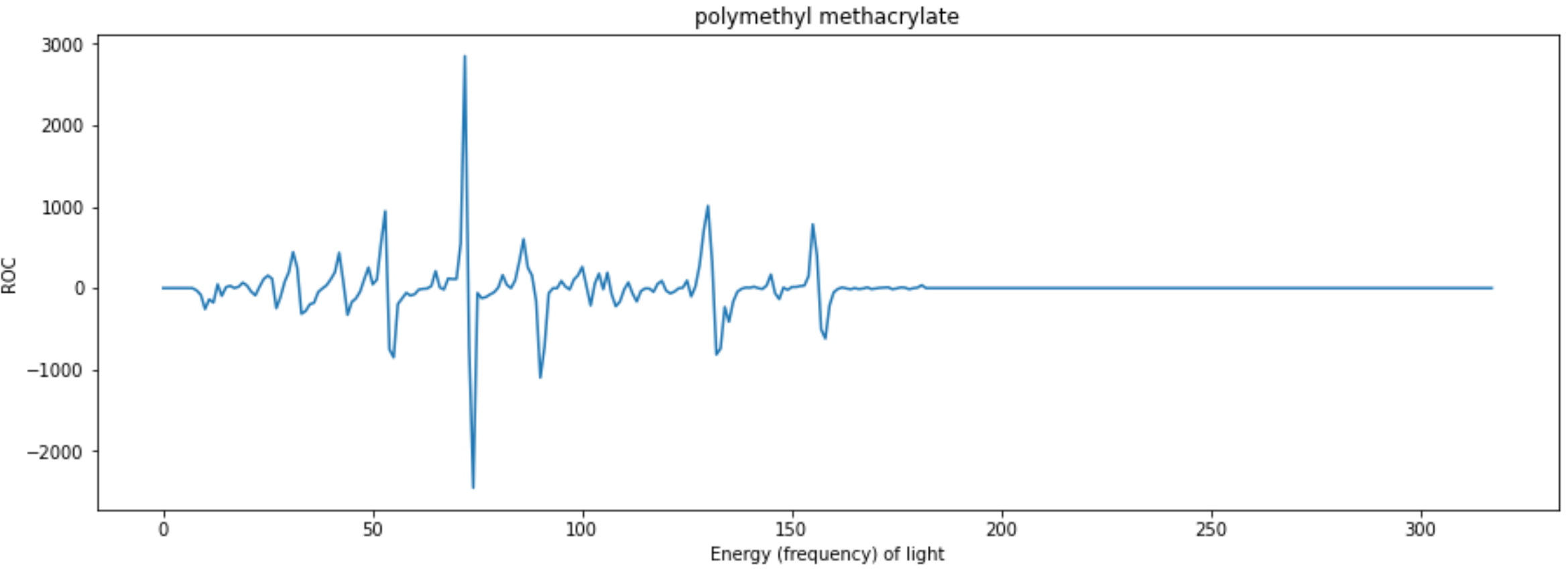}
	\centering
	\caption{Binning technique of ROC processed sample for polymethyl methacrylate.}
	\label{fig:roc_binning}
\end{figure}

Since Raman spectroscopy data has the characteristics of time-series data, the spikes in the distribution are the most important patterns that can be extracted from the samples. An equal-width binning technique was used in this research.  Then a smoothing by bin-means technique is applied where the average of the values in a bin is calculated and each bin is now represented using the average value. Fig. \ref{fig:roc_binning} shows the results of this technique applied to a single sample for polymethyl methacrylate type with a bin width of 11. That is, every 11 values are mapped to the same bin, and the average value  of the bin is calculated.  One can also observe the compressed scale on the x-axis as compared to the scale in Fig.~\ref{fig:roc_changed}.

\subsection{Augmentation}
\label{sec:augment}

As the dataset is very small, the data augmentation function has been implemented to populate the training dataset with more samples. The pseudo-code for the augmentation process is given in Algorithms~\ref{alg:augmentation}-\ref{alg:get_full_augmented_example}.

The augmentation function works the following way: firstly, the function iterates over a polymer (plastic) type that needs to be augmented, and the \textcolor{blue}{pct\_change} function is used to calculate the change between the current and the previous value of a sample, by dividing the two numbers.  This helps to keep information about the changing values. 

Secondly, a random uniform distribution is applied, where random values between -0.05 and 0.05 are chosen.  This is a user-defined parameter \textcolor{blue}{random\_change} and controls how much the augmented dataset differs from the original sample.
 
The last step is to reverse the percentage change function, by multiplying the original value with the new percentage change value. As the percentage change value has been changed slightly, each generated value is different from the original value. However, such a change leads to a problem of rapidly increasing or decreasing graph fluctuations. These sharp fluctuations are controlled by the \textcolor{blue}{max\_pct\_change} parameter.  This parameter value is set to 99, meaning that the generated value could be up to 99\% more than the original value or 99\% less than the original value.  Additionally, Alg.~\ref{alg:augmentation} includes parameter \textcolor{blue}{shift}.  This parameter is meant to shift the values on the y-axis (higher or lower). However, this value was set to 0, as it does not change the test accuracy significantly. 

\begin{algorithm}
\caption{Generate\_Augmented\_Data(train\_dataset, plastic\_type\_list, min\_num\_examples, random\_change=0.05, shift=0, max\_pct\_change=99)}\label{alg:augmentation}
\begin{algorithmic}[1]

\State $plastic\_type\_list \gets [el.lower()$ for $el$ in $plastic\_type\_list]$
\State $train\_dataset\_augm \gets train\_dataset$

\For{$plastic\_type$, $idxs$ in $train\_dataset.items()$}

    \State $iterate \gets 0$
    
     \If{$plastic\_type$ in $plastic\_type\_list$} 
        \While{$len(train\_dataset\_augm[plastic\_type])$ $<$ $min\_num\_examples$}
            \State $cur\_idx \gets iterate$ \% $len(idxs)$
            \State $pct\_change\_list \gets pct\_change(idx[cur\_idx]['y'])$
            
            \State $augm\_example \gets get\_augmented\_example(pct\_change\_list, random\_change)$
            \State $init\_value \gets idxs[cur\_idx]['y'][0]$
            
            \If{$min(idxs[cur\_idx]['y']) \leq 0$} 
                \State $init\_value \gets init\_value + abs(min(idxs[cur\_idx]['y'])) + 1$
            \EndIf
            
            \State $augmented\_data \gets get\_full\_augmented\_example(idxs[cur\_idx]['y'],$ $augm\_example,$ $init\_value,$ $shift,$ $max\_pct\_change)$
            
            \State $train\_dataset\_augm[plastic\_type] \gets train\_dataset\_augm[plastic\_type] + DataFrame(\{'x': idxs[cur\_idx]['x'],$ $'y': augmented\_data\})$
            \State $iterate \gets iterate + 1$
        \EndWhile
    \EndIf
    
\EndFor

\State $train\_dataset \gets train\_dataset\_augm$

\end{algorithmic}
\end{algorithm}

\begin{algorithm}
\caption{pct\_change(dataframe)}\label{alg:augmentation}
\begin{algorithmic}[1]

\State $pct\_change\_list \gets []$
\State $min\_value \gets min(dataframe)$

\If{$min\_value \leq 0$} 
    \State $dataframe \gets [i + abs(min\_value) + 1$ for $i$ in $dataframe]$
\EndIf

\For{$num$, $\_$ in $enumerate(dataframe[$:$len(dataframe) - 1])$}

    \State $pct\_change\_list \gets pct\_change\_list + (dataframe[num + 1] / dataframe[num])$
    
\EndFor

\State return $pct\_change\_list$

\end{algorithmic}
\end{algorithm}

\begin{algorithm}
\caption{get\_augmented\_example(pct\_change\_list, random\_change=0.2)}\label{alg:get_augmented_example}
\begin{algorithmic}[1]

\State $augm\_pct\_change\_list \gets []$

\For{$el$ in $pct\_change\_list$}
    \State $tmp\_el \gets el + random.uniform(-random\_change, random\_change)$
    
    \If{$tmp\_el \leq 0$} 
        \State $tmp\_el \gets el$ \Comment{el is $>$ 0, because of $pct\_change$ function}
    \EndIf
    
    \State $augm\_pct\_change\_list \gets augm\_pct\_change\_list + tmp\_el$
\EndFor

\State return $augm\_pct\_change\_list$

\end{algorithmic}
\end{algorithm}

\begin{algorithm}
\caption{get\_full\_augmented\_example(original\_dataset, pct\_change\_list, init, shift=0, max\_pct\_change=10)}\label{alg:get_full_augmented_example}
\begin{algorithmic}[1]

\State $previous\_value \gets init + shift$
\State $augm\_pct\_change\_list \gets []$
\State $augm\_pct\_change\_list \gets augm\_pct\_change\_list + previous\_value$

\State $min\_value \gets min(original\_dataset)$

\If{$min\_value \leq 0$} 
    \State $original\_dataset \gets [i + abs(min\_value) + 1$ for $i$ in $original\_dataset]$
\EndIf

\For{$num$, $el$ in $enumerate(pct\_change\_list)$}
    \State $previous\_value \gets previous\_value * el$
    
    \If{$previous\_value > original\_dataset[num + 1] * (1 + max\_pct\_change / 100)$} 
        \State $previous\_value \gets original\_dataset[num + 1] * (1 + max\_pct\_change / 100)$
    \EndIf
    
    \If{$previous\_value < original\_dataset[num + 1] * (1 - max\_pct\_change / 100)$} 
        \State $previous\_value \gets original\_dataset[num + 1] * (1 - max\_pct\_change / 100)$
    \EndIf
    
    \State $augm\_pct\_change\_list \gets augm\_pct\_change\_list + previous\_value$
\EndFor

\State return $augm\_pct\_change\_list$

\end{algorithmic}
\end{algorithm}

\begin{figure}[h!]
	\includegraphics[width=12cm]{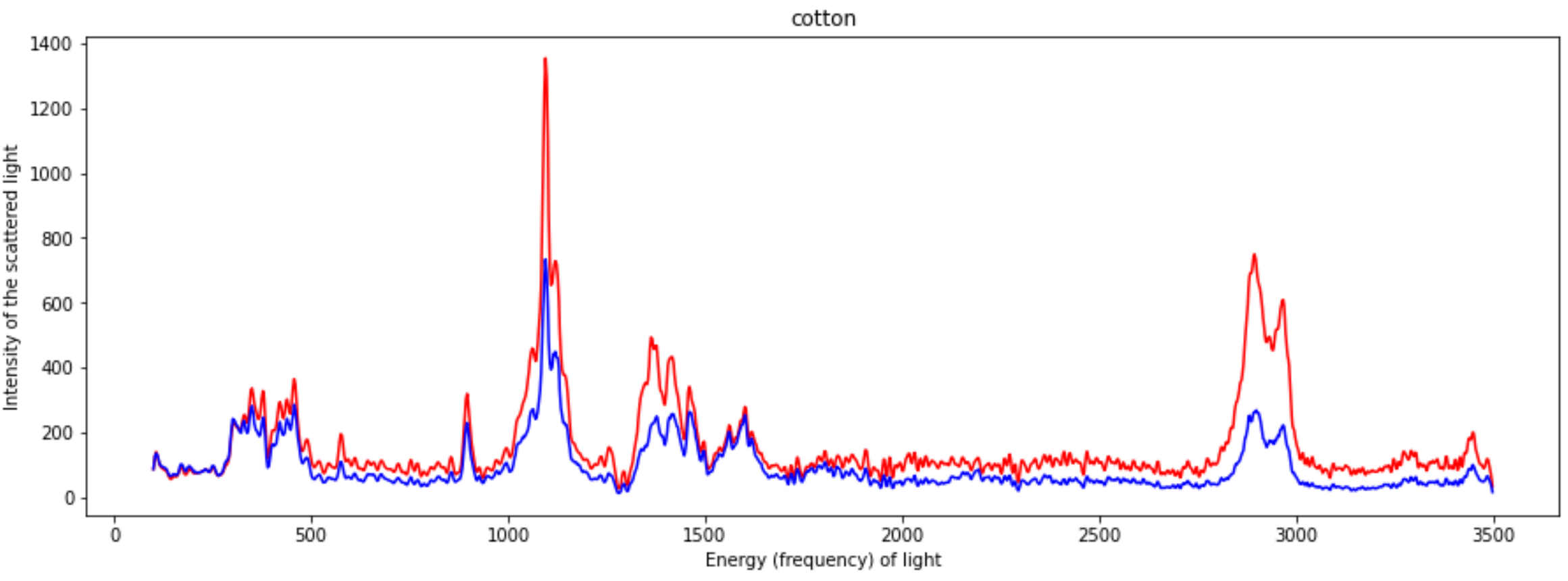}
	\centering
	\caption{Improved augmented data (red – original; blue – augmented).}
	\label{fig:augmented_example}
\end{figure}

Example of augmented data is shown in Fig. \ref{fig:augmented_example}. The line which is coloured red is the original sample, and a blue line is the augmented sample. It can be seen, that the augmented sample keeps the same trajectory as the original sample, but introduces some changes on the y-axis values. Spikes on generated samples are retained on same x-axis values, however, the intensity of such spikes is different.

In our research,  we have augmented polymer types that have either a small number of samples (e.g., less than 5) or have performed poorly on test results.

\section{RESULTS AND DISCUSSION}
\label{sec:res}

In this section, we analyze the results of the experiments. The following ML algorithms have been used in this research using the scikit-learn workbench\footnote{https://scikit-learn.org/stable/}:  support vector machines (SVM), random forest (RF), decision tree (DT), k-nearest neighbours (KNN), and artificial neural network (ANN). The RF model is the only model that gives high classification accuracy.  Hence, in our discussions related to the analysis of the effect of different preprocessing, discretization as well as augmentation techniques, we will use the RF model as our baseline model.   

In an effort to increase the training set size, we also experimented with another microplastic dataset \textcolor{blue} {Open Specy} \footnote{https://doi.org/10.1021/acs.analchem.1c00123.s001}.  This dataset contains a total of 183 examples and 137 polymer types~\cite{Cowger2021}. We observed that most of the polymer types in this dataset do not appear in SLoPP, therefore, cannot be used. Additionally, the intensities of the scattered light (y-axis) for Open Spacy dataset is normalized to values between 0 and 1, and there the original values for the intensities cannot be reconstructed. As a result, this dataset was not used in our final model training experiments.

\begin{center}
\begin{table}[h!]
\centering
		\begin{tabular}{|l|l|l|}
		\hline
        \textcolor{blue}{Experiment}                  & \textcolor{blue}{Preprocessing Methods}                  & \textcolor{blue}{RF  Accuracy (\%)} \\
				\hline
        1 & scaling (x-axis), ROC       & \textcolor{blue}{79.38}                  \\
        2 & scaling (x-axis), no ROC    & 61.85                  \\
        3 &  no scaling (x-axis), ROC    & 72.16                  \\
        4 &  no scaling(x-axis), no ROC & 53.61       \\
				\hline
    \end{tabular}
    \caption{Model accuracy with different variations of preprocessing functions.}
    \label{tab:results_roc_scale_bin}
\end{table}
\end{center}

Table~\ref{tab:results_roc_scale_bin}  presents experiments using different preprocessing functions and with no scaling of the y-axis values.  All experiments were conducted with a combination of the scaling and ROC transformation methods described in sections~\ref{sec:norm} and ~\ref{sec:roc}. The best result (accuracy of 79.38\% highlighted in blue) was achieved with scaling energy values and using the rate of change transformed feature.

\begin{figure}[h!]
	\includegraphics[width=12cm]{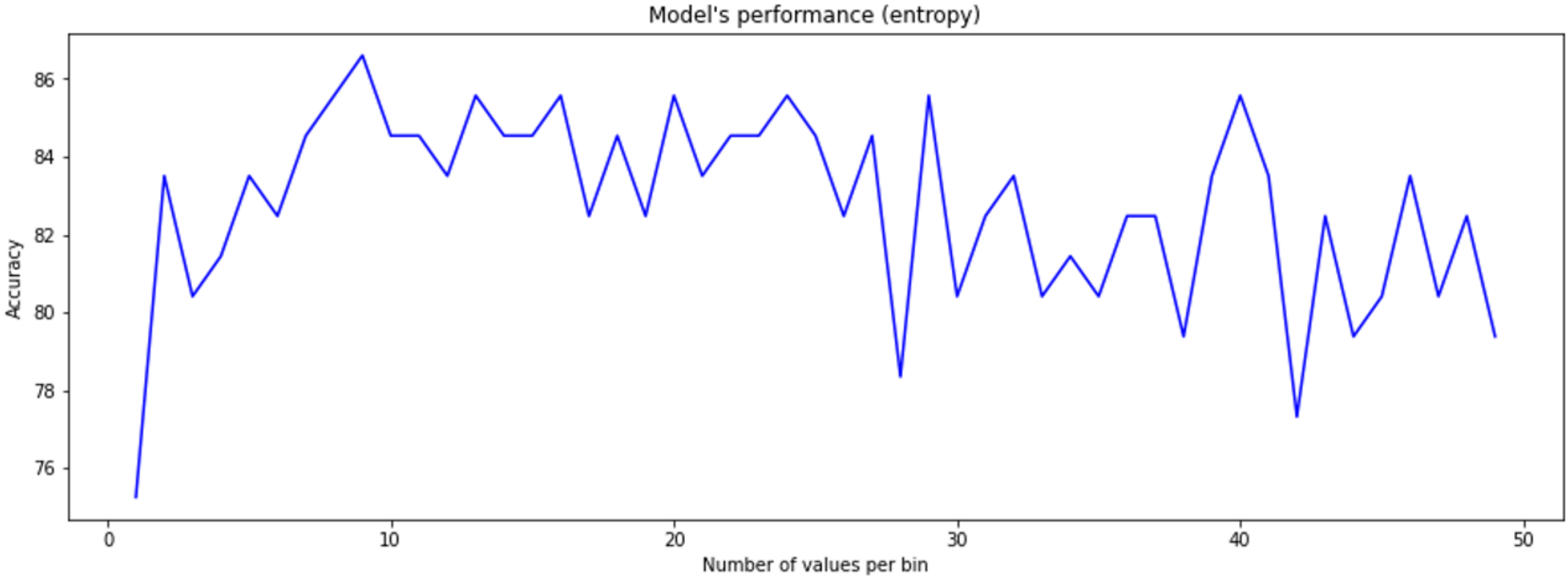}
	\centering
	\caption{Accuracy of the model with different bin sizes.}
	\label{fig:results_binning_no_augm}
\end{figure}

Fig.~\ref{fig:results_binning_no_augm} shows the performance of the RF model using different bin sizes ranging from 2 to 50.  The discretization technique which achieves the best result (classification accuracy 86.59\% with information gain criteria) is when the bin size is between 10 and 20. This experiment does not use any augmentation method.  

\begin{figure}[h!]
	\includegraphics[width=12cm]{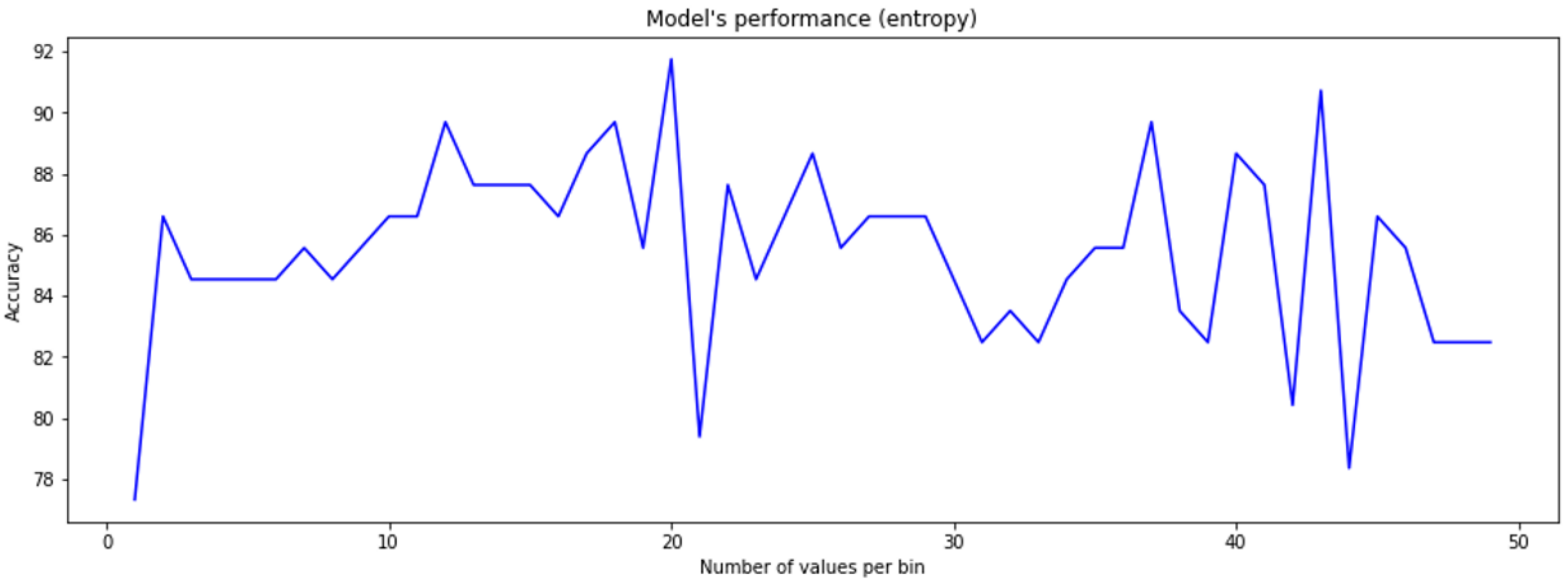}
	\centering
	\caption{Accuracy of the model with different bin size with augmented data with \textcolor{blue}{information gain} criteria.}
	\label{fig:results_binning_augm}
\end{figure}

\begin{figure}[h!]
	\includegraphics[width=12cm]{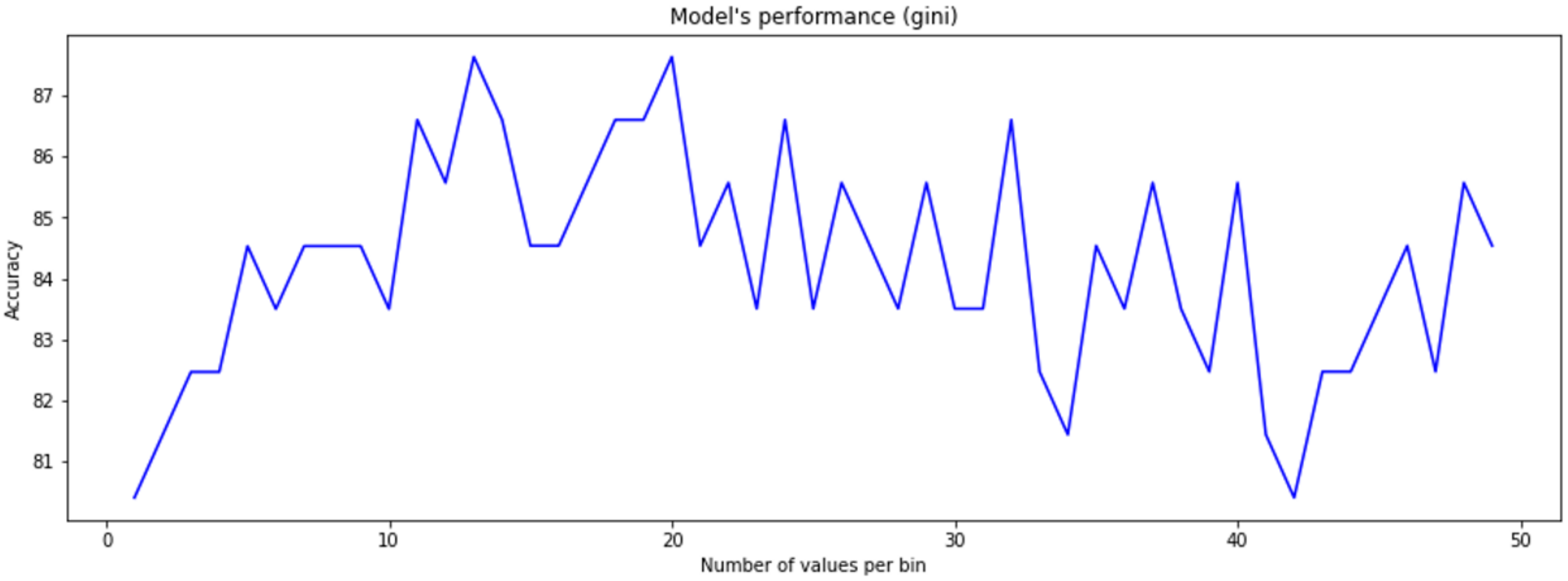}
	\centering
	\caption{Accuracy of the model with different bin size with augmented data with \textcolor{blue}{gini} criteria.}
	\label{fig:results_binning_augm_gini}
\end{figure}

However, if the  training dataset is augmented, the accuracy increases dramatically (see Fig. \ref{fig:results_binning_augm}). Augmentation has been applied to  the following polymer types: Cellulose Acetate, Polyamide, Polymethyl Methacrylate and Polyurethane, where each type has been augmented up to 15 examples.   Experiments were conducted with two different criteria for the RF model, information gain (in Fig.~\ref{fig:results_binning_augm}) and gini (in Fig.~\ref{fig:results_binning_augm_gini}). One can observe that the classification accuracy is not as good with the gini criteria as with the information gain criteria.

The best result of 91.75\% classification accuracy was obtained with a bin size of 12 and information gain (entropy) as the criteria for tree construction.

\begin{figure}[h!]
	\includegraphics[width=12cm]{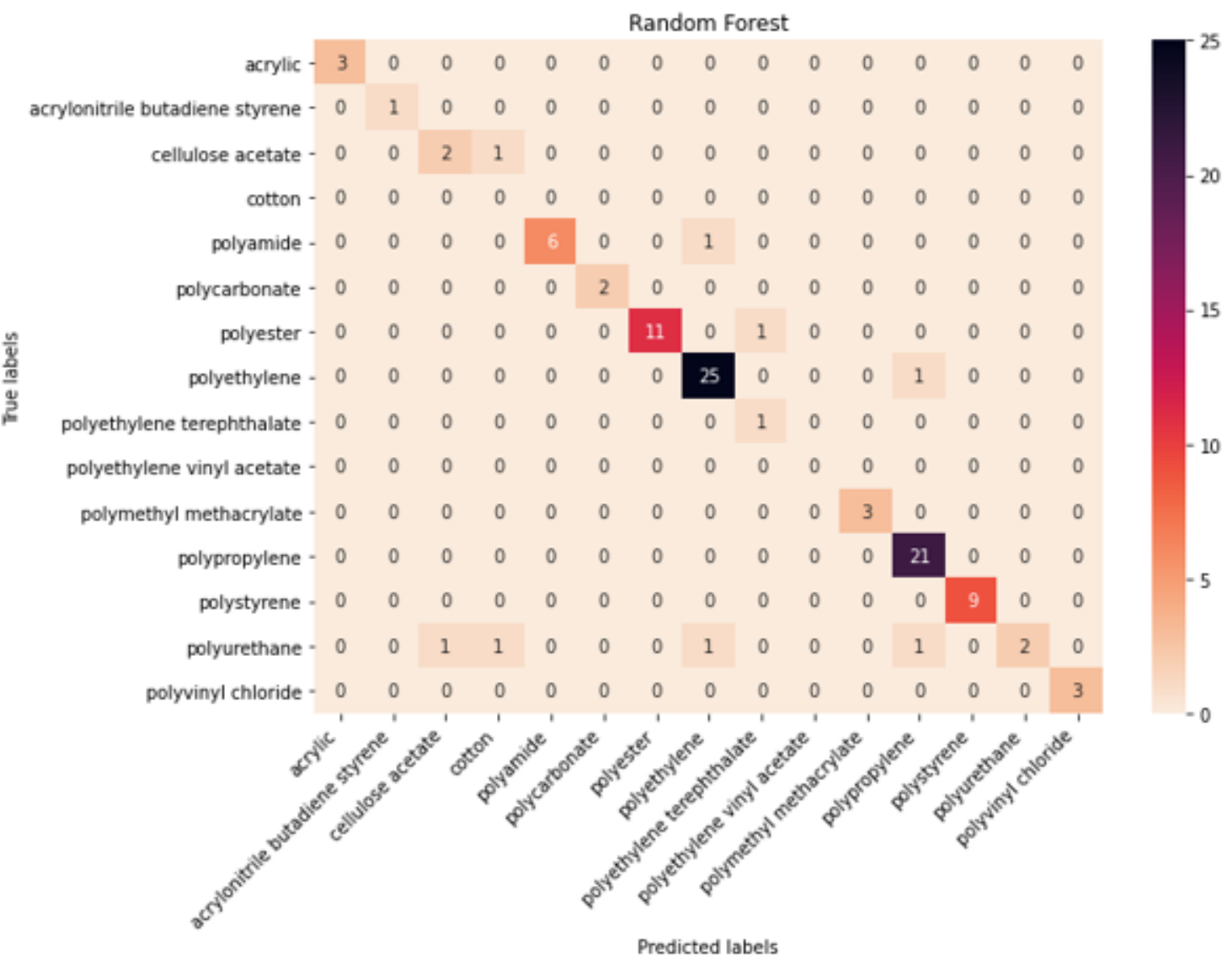}
	\centering
	\caption{Confusion matrix for the model that achieved 91.75\% accuracy.}
	\label{fig:conf_mat_91}
\end{figure}

Fig. \ref{fig:conf_mat_91} gives the confusion matrix with classification details for each polymer type in the SLOPP-E dataset. It can be seen, that the model detects most of the samples correctly. However, it misclassifies a few samples, especially the  Polyurethane polymer where 4 out of 6 samples were misclassified. The training accuracy of this model is 100\%, which signifies that the model overfits.  This is due to the fact that that the model was trained on just 306 samples.

\begin{figure}[h!]
	\includegraphics[width=12cm]{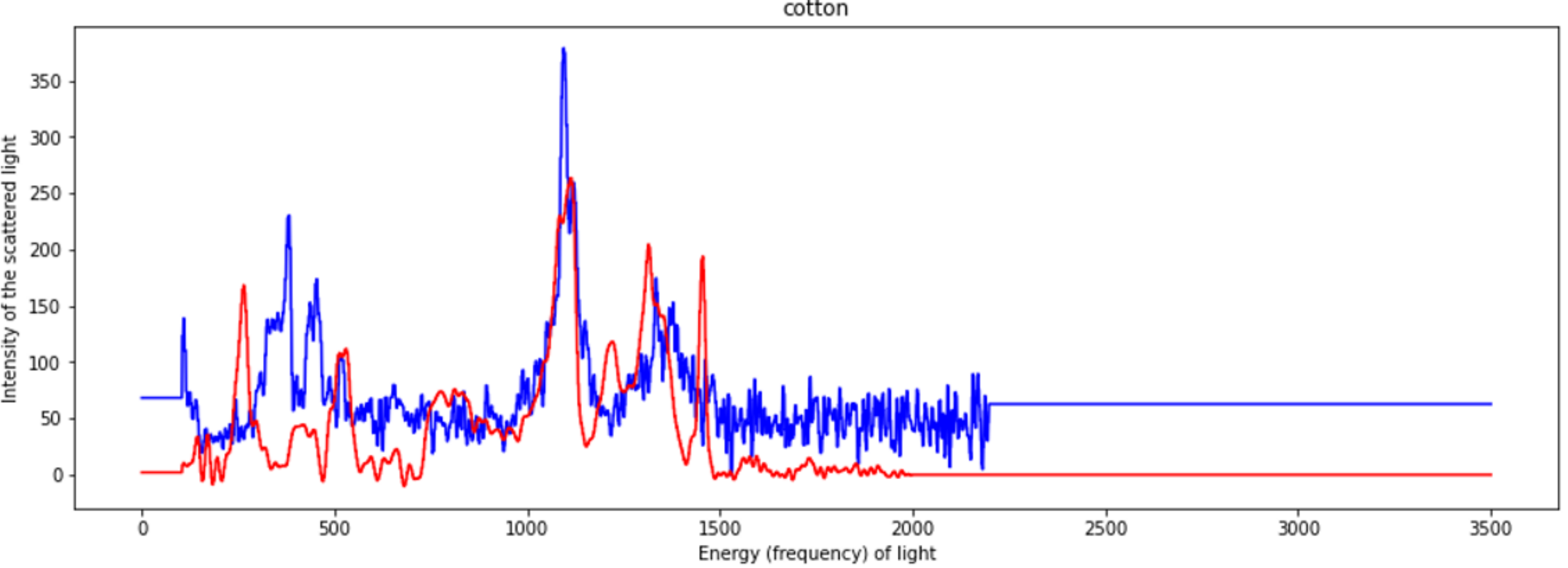}
	\centering
	\caption{Wrongly detected sample (red) plotted on a wrongly predicted type (blue).}
	\label{fig:results_wrong_detected_1}
\end{figure}

\begin{figure}[h!]
	\includegraphics[width=12cm]{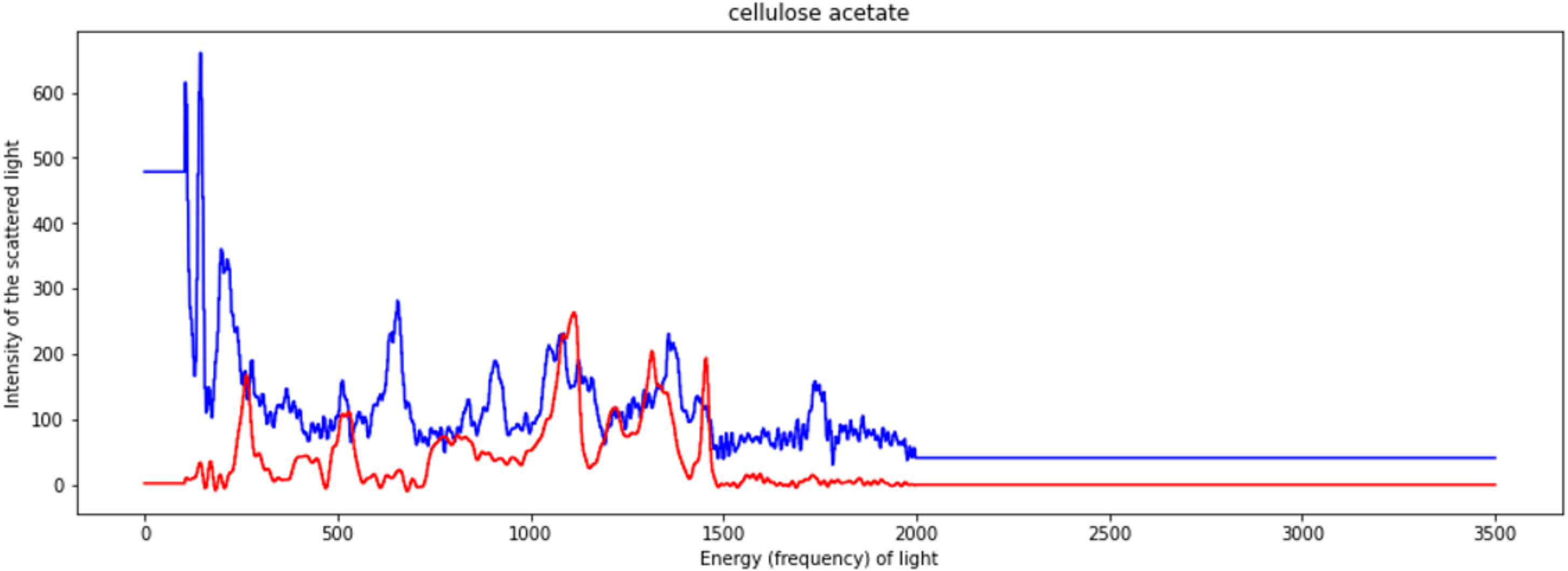}
	\centering
	\caption{Wrongly detected sample (red) plotted on a correct type (blue).}
	\label{fig:results_wrong_detected_2}
\end{figure}

Table~\ref{tab:misclassified_cases} shows the most commonly occurring misclassified samples.   The model always predicts the same polymer type for these samples, irrespective of how the data has been processed.  Upon examination of the corresponding Raman spectroscopy plots, it is hard to detect whether the sample is mislabeled or the model predicts the result wrongly (Fig. \ref{fig:results_wrong_detected_1}-\ref{fig:results_wrong_detected_2}). Fig. \ref{fig:results_wrong_detected_1} shows a sample of a type cellulose acetate, plotted with one example from a train dataset of cotton type (incorrect type). The spike around 1000 value on the x-axis has the same shape as other spikes  which do not fully correspond to this type (cotton). Fig. \ref{fig:results_wrong_detected_2} shows a sample of cellulose acetate type, plotted with one example from a train dataset of the same type (correct type). It can be observed, that these samples have a different shape compared to the one in Fig. \ref{fig:results_wrong_detected_1}.  Hence  these samples do not match. 

\begin{center}
\begin{table}[h!]
\centering
		\begin{tabular}{|l|l|l|}
		\hline
       \textcolor{blue}{Sample number} & \textcolor{blue}{Model predicted}   & \textcolor{blue}{Actual label} \\
			\hline
        6                      & Cotton                     & Cellulose Acetate     \\
        12                     & Polyethylene               & Polyamide             \\
        24                     & Polyethylene Terephthalate & Polyester             \\
        50                     & Polyurethane               & Polyethylene          \\
        88                     & Cellulose Acetate          & Polyurethane          \\
        89                     & Polyamide                  & Polyurethane  \\
			\hline
    \end{tabular}
    \caption{Misclassified cases.}
    \label{tab:misclassified_cases}
\end{table}
\end{center}

\begin{figure}[h!]
	\includegraphics[width=12cm]{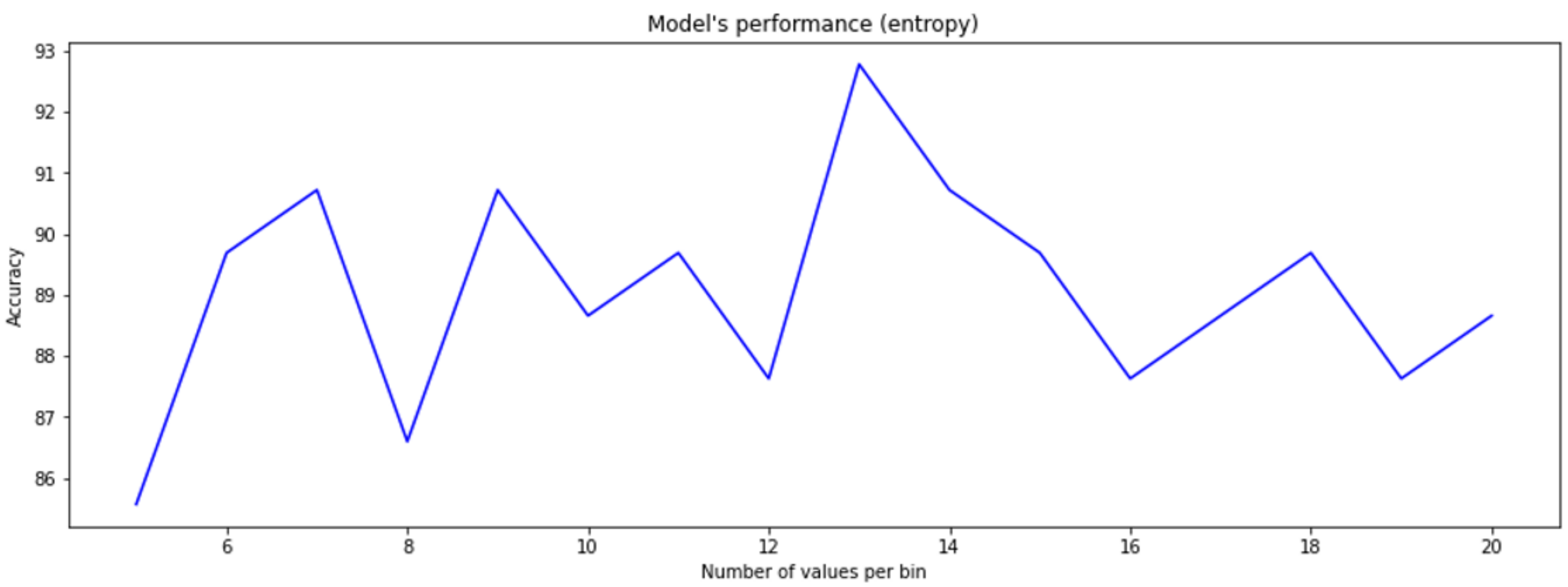}
	\centering
	\caption{Accuracy of the model with noise added to the training dataset.}
	\label{fig:results_noise}
\end{figure}

Since the SLoPP-E test set was subject to weather and ageing, another experiment was conducted by adding some noise to the training (SLoPP) dataset to introduce some non-linearity. For each value on the x-axis, a small random value was either added or subtracted. However, the addition of noise did not change the accuracy in any significant way (see Fig. \ref{fig:results_noise}).

The final model was trained on augmented data, which was preprocessed using the following functions: ROC, scaling x-axis (0-3500), discretization with the window size 12, no y-axis rescaling. The following polymer types were augmented: Cellulose Acetate, Polyamide and Polyurethane (30 samples), Polyester (40 samples), Polymethyl Methacrylate (10 samples) and Polystyrene (20 samples). 

\begin{figure}[h!]
	\includegraphics[width=12cm]{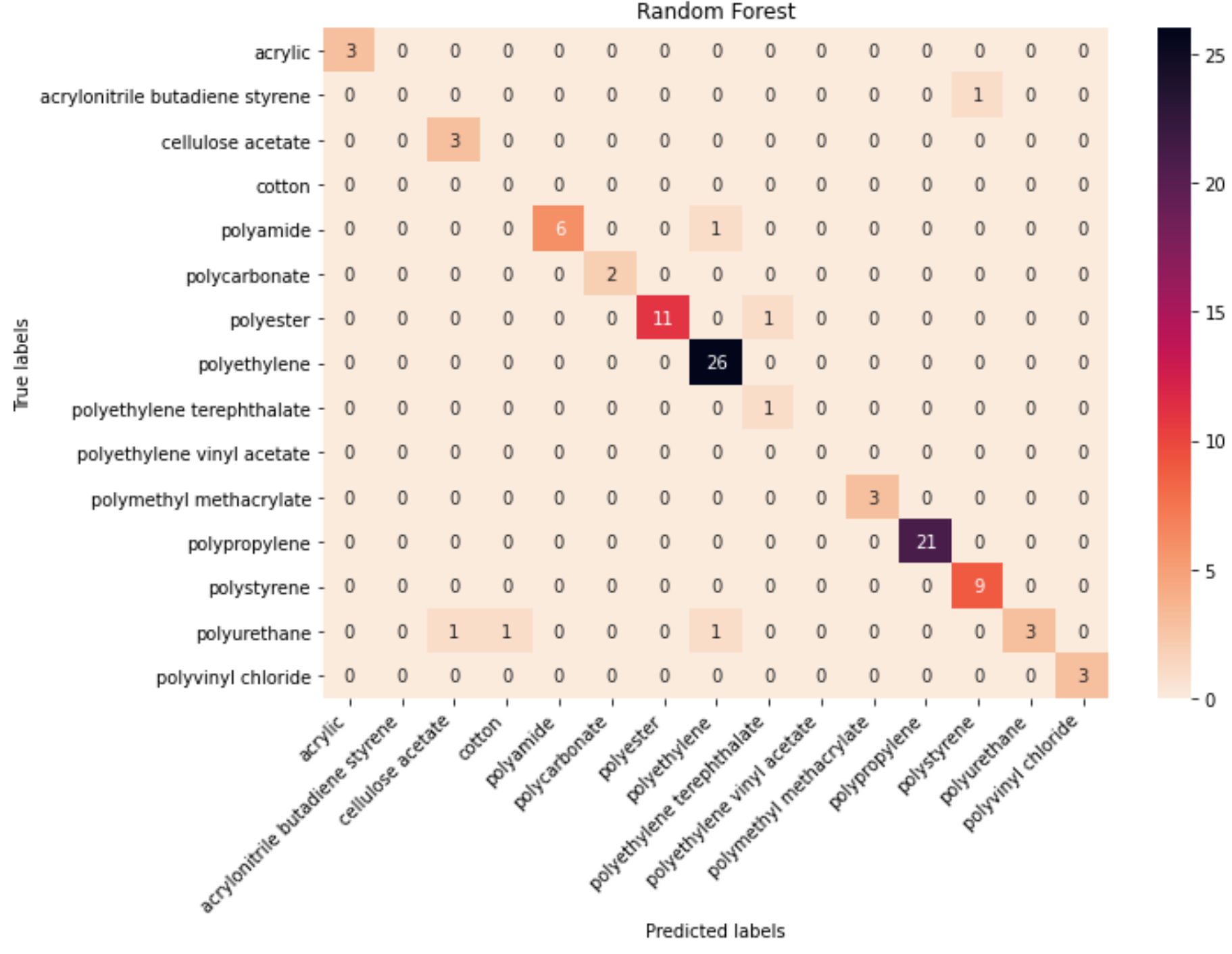}
	\centering
	\caption{Confusion matrix for the model that achieved 93.81\% accuracy.}
	\label{fig:conf_mat_final}
\end{figure}

Figure~\ref{fig:conf_mat_final} gives the confusion matrix where the trained model detects most of the samples correctly, and only a few samples are mislabeled. The model mostly performs poorly on the Polyurethane type, as it misclassifies 3 out of 6 test samples. The Acrylonitrile Butadiene Styrene type also performs poorly, as it misclassifies a single test sample. Since there is only 1 test sample available, this classification could be misleading.

Table~\ref{tab:results_ml_models} gives the results of experiments with other models.  However, none of the other models achiev the same accuracy on the test dataset as the random forest model. The ANN model with 4 layers of 128, 64, 32 and sparse categorical cross-entropy was used with the adam optimizer. SVM with linear kernel (it produced the best accuracy, compared to other kernels), DT (with entropy)  and KNN with 3 nearest neighbours were used.

\begin{center}
\begin{table}[h!]
\centering
		\begin{tabular}{|l|l|}
		\hline
        \textcolor{blue}{Models}          & \textcolor{blue}{Classification Accuracy} \\
				\hline
        ANN                     & 71.13\%           \\
        SVM (linear kernel)     & 73.19\%           \\
        DT  & 69.07\%           \\
        KNN                     & 73.19\%   \\
							\hline
    \end{tabular}
    \caption{Accuracies of different machine learning models.}
    \label{tab:results_ml_models}
\end{table}
\end{center}

In summary, our experiments demonstrate that there is a significant improvement in the classification accuracy (from 89\% to 93.81\%) when the dataset is augmented.  This shows that a larger data set with more training and balanced samples can improve the classification performance beyond 94\% and learn from environmentally degraded samples.  The other important issue is that there is some concern that the original sample maybe mislabeled.  This is because the predicted type (by the model) is not similar to the actual type (visually). Another observation is that even when wave-like samples (from the Mendeley dataset) were excluded from the training set, the classification accuracy was around 90\%.  This shows that adding the samples (even though some of the shapes were different) may have in fact helped the model to learn,  or at least, did not have a negative effect on the model.  One reason could be that SLoPP-E (test dataset) does not have similar wave-like samples.

\section{Conclusion}
\label{sec:con}
In this work, we were primarily interested in detecting polymer types from the spectral signature of Raman spectroscopy microplastics data which were environmentally aged from a well-known dataset.  Environmental weathering occurs from exposure to temperature extremes, UV radiation, wind, water erosion in freshwater environments, and saltwater erosion in marine environments, in addition to other factors in localized ecosystem contexts.  Exposure of microplastics to the environment affects their spectrographic output data, making spectrographic analysis results less reliable than unaffected samples.  Different normalization methods as well as data transformation methods for preprocessing and feature engineering were applied.  Since the number of training samples in certain polymer types were limited, a data augmentation method was used. Different ML models were trained with the random forest model giving the best result with an improvement in classification accuracy of 93.81\% from 89\%. A detailed discussion of the results is presented in an effort to contribute to the understanding chemical compounds of plastics that have been weathered by various environmental processes.   The significance of this research project is to strive for a measurably improved predictive capacity of Raman spectroscopy data to help classify polymer types through an applied machine learning process. This work can lead to applications in ecotoxicology and environmental research, the circular economy for plastics recycling processes, water quality testing and treatment processes, food and beverage quality control testing, to name a few.

\bibliographystyle{elsarticle-num}
\bibliography{./Raman}

\end{document}